\documentclass[conference]{IEEEtran}
\usepackage{cite}
\usepackage{amsmath,amssymb,amsfonts}
\usepackage{graphicx}
\usepackage{textcomp}
\usepackage{algorithm}
\usepackage{algpseudocode}
\usepackage{booktabs}
\usepackage{subcaption}
\usepackage{placeins}
\graphicspath{{figures/}}
\usepackage{hyperref}

\begin{document}

\title{SNGR: Selective Non-Gaussian Refinement\\
for Ambiguous SLAM Factor Graphs}

\author{
\IEEEauthorblockN{Anushka Kulkarni, Sarthak Dubey}
\IEEEauthorblockA{
  \textit{Northeastern University, Boston, MA, USA}\\
  kulkarni.anushka@northeastern.edu, dubey.sart@northeastern.edu}}

\maketitle

\begin{abstract}
We present Selective Non-Gaussian Refinement (SNGR), a SLAM framework
that augments iSAM2 with targeted nested sampling on windows where
Gaussian approximations are likely to fail.
We detect such regions using the condition number of joint marginal
covariances and selectively refine them using the full nonlinear
factor graph likelihood, with a gating mechanism to avoid degradation
in multimodal cases.
Experiments on range-only SLAM with wrong data association show that
SNGR achieves high-precision failure detection and consistent local
likelihood improvements while reducing computational cost relative
to exhaustive non-Gaussian inference.
These results highlight both the promise and the limitations of
selective refinement for approximate SLAM posteriors.
\end{abstract}


\section{Introduction}

Simultaneous localization and mapping (SLAM) estimates a robot's
trajectory and environment map from sensor measurements.
The leading formulation represents the joint posterior as a factor
graph and solves for the MAP estimate by minimizing squared
Mahalanobis errors~\cite{dellaert2017factor}.
Incremental solvers such as iSAM2~\cite{kaess2012isam2} maintain a
Bayes tree that re-linearizes only affected cliques when new
measurements arrive, enabling real-time deployment.

These methods rely on a Gaussian approximation of the posterior.
When this assumption breaks, most notably under range-only sensing
and wrong data association, the MAP estimate can be both incorrect
and overconfident.
Under range-only sensing, a measurement $r=\left\|x_t-l_k\right\|_2$
constrains the robot to a circle; when a landmark is observed from
similar directions, the posterior is bimodal and iSAM2 converges to
the saddle between modes.
Under wrong data association, an inconsistent factor biases the MAP
while the covariance remains small, producing a confident but
incorrect estimate.

Huang, Papalia, and Leonard~\cite{huang2022nested} showed nested
sampling accurately characterizes non-Gaussian SLAM posteriors, but
position it as an offline reference tool unsuitable for online
deployment.

Applying nested sampling to every window is computationally
unjustified when most regions of the factor graph are well-conditioned and their Gaussian
posteriors are accurate.
We adapt the condition-number degeneracy analysis of
Tuna et al.~\cite{tuna2024informed} from point-cloud registration
to the iSAM2 marginal covariance: a high condition number indicates
a poorly conditioned local landscape where the linearized solver is
likely to fail and nested sampling may recover a better estimate.
Windows exceeding threshold $\tau$ receive nested sampling via
dynesty~\cite{speagle2020dynesty}; others retain iSAM2 estimates.

There are two key aspects to our work: the use of factor graphs to improve the underlying inference algorithm and the use of nested sampling to perform approximate stochastic inference. We will first survey inference techniques in SLAM with a focus on those leveraging factor graph structures. We will then discuss general stochastic inference techniques which have not yet been applied to SLAM. The contributions of this work are a condition-number trigger for detecting non-Gaussian failure regions in SLAM factor graphs, a selective nested sampling framework compatible with iSAM2, and an empirical analysis of when selective non-Gaussian inference succeeds and fails.

All code and experimental results are available at \url{https://github.com/anushkakulk/sngr-slam}.

\section{Related Work}

\subsection{Deterministic SLAM Inference}

iSAM~\cite{kaess2008isam} introduced incremental updates to avoid
refactorization of the SLAM factor graph; iSAM2~\cite{kaess2012isam2} refined this with the
Bayes tree and fluid relinearization.
The g2o framework~\cite{kummerle2011g2o} offers an alternative
solver~\cite{cadena2016past}.

Switchable constraints~\cite{sunderhauf2012towards}, dynamic
covariance scaling~\cite{agarwal2013robust}, and the max-mixture
model~\cite{olson2013inference} add robustness to wrong associations
but produce a single robust MAP rather than the full posterior.

\subsection{Stochastic Non-Gaussian SLAM}

Particle filter SLAM~\cite{thrun2005probabilistic} tracks arbitrary
distributions but scales exponentially with state dimension.
Rao-Blackwellized filtering~\cite{montemerlo2002fastslam} reduces
dimensionality but landmark count limits scalability.

\subsection{Nested Sampling for SLAM}

Nested sampling maintains live points contracting toward
higher-likelihood regions, terminating when evidence converges.
Unlike MCMC, it explores multimodal distributions without chains
becoming stuck.
Speagle~\cite{speagle2020dynesty} developed dynesty, which our
implementation uses.

NSFG~\cite{huang2022nested} splits the factor graph into prior and
likelihood sets, draws samples via ancestral sampling, and outperforms
NUTS and SMC by over an order of magnitude.
SNGR differs: we use the iSAM2 MAP as sampling prior and apply
sampling selectively on triggered windows.

\subsection{Degeneracy Detection}
Tuna et al.~\cite{tuna2024informed} showed that $\lambda_{\max}/\lambda_{\min}$
of the registration Hessian identifies weakly constrained directions
in point-cloud alignment.
The Hessian of the negative log-posterior evaluated at the MAP is
the inverse of the Gaussian approximation covariance, so a
high condition number on the Hessian is equivalent to a high
condition number on the marginal covariance.
Both indicate an elongated, pencil-shaped uncertainty ellipsoid
where the Gaussian approximation is most likely to be a poor
representation of the true posterior: the flat directions of the
cost surface are precisely where nonlinear curvature, secondary
modes, and outlier-induced bias accumulate.
 
We adapt this insight to the SLAM setting by operating on the
iSAM2 marginal covariance rather than the registration Hessian.
This is advantageous because iSAM2 maintains the Bayes tree
factorisation of the joint covariance, making per-window marginals
available at low cost via back-substitution without recomputing the
full Hessian after each update.
A window whose marginal covariance has a high condition number is
therefore one where the Gaussian solver is operating in a poorly
conditioned landscape, and where targeted non-Gaussian refinement
is likely to recover a better estimate.

\section{Methods}

\subsection{Problem Formulation}

Let $x_t\in SE(2)$ denote robot pose and $l_k\in\mathbb{R}^2$
landmark $k$.
Joint state $\Theta=\{x_0,\ldots,x_{T-1},l_0,\ldots,l_{K-1}\}$ is
encoded in factor graph $\mathcal{G}$ with: prior on $x_0$; odometry
factors $f_t^{\text{odom}}(x_{t-1},x_t)=
\mathcal{N}(x_t;\,x_{t-1}\oplus u_t,\Sigma_u)$; range factors
$f_{t,k}(x_t,l_k)=\mathcal{N}(z_{t,k};\,\left\|x_t^{\text{pos}}
-l_k\right\|_2,\sigma_r^2)$; and weak landmark priors after three
observations.

The joint posterior relates to all factors as:
\begin{equation}
  p(\Theta | \mathcal{Z}) \propto \prod_f f(\Theta_f)
  \label{eq:posterior}
\end{equation}
and iSAM2 finds the MAP estimate:
\begin{equation}
  \hat\Theta = \arg\min_\Theta \sum_f
  \|h_f(\Theta) - z_f\|^2_{\Sigma_f}
  \label{eq:map}
\end{equation}
where $h_f$ is the measurement function of factor $f$ and
$\|\cdot\|_{\Sigma_f}$ is the Mahalanobis norm.

Wrong-association noise corrupts fraction $p_\text{noise}$ of
measurements by replacing correct index $k$ with incorrect $j\neq k$.
The factor is geometrically irreconcilable but iSAM2 has no rejection
mechanism.

For window $\mathbf{w}(t)=\{x_t,x_{t+1},x_{t+2}\}$, ground-truth
failure is defined at $\varepsilon=0.5$\,m:
\begin{equation}
  \phi(\mathbf{w}) = \mathbf{1}\!\left[
  \tfrac{1}{3}\sum_{s=t}^{t+2}
  \left\|\hat{x}_s^{[x,y]}-x_s^{\text{gt}}\right\|_2>\varepsilon
  \right]
  \label{eq:fail}
\end{equation}

\subsection{Condition-Number Trigger}

For each window:
\begin{equation}
  s_\mathbf{w} = \log_{10}\!\left(
    \lambda_{\max}(\Sigma_\mathbf{w})\,/\,
    \lambda_{\min}(\Sigma_\mathbf{w})
  \right)
  \label{eq:score}
\end{equation}
where $\Sigma_\mathbf{w}\in\mathbb{R}^{9\times9}$ is the joint
marginal covariance.
Large $s_\mathbf{w}$ signals elongated uncertainty: one direction
tightly constrained, the orthogonal nearly unconstrained.

Window $\mathbf{w}$ triggers when $s_\mathbf{w}>\tau$ and $t\geq W-1$.
Threshold $\tau$ is calibrated by selecting the largest value
producing zero triggers at $p=0.0$ while maximizing recall at
$p=0.3$.

\subsection{Variable Closure and Sampling}

Closure $\mathcal{C}_\mathbf{w}$ includes all variables in factors
connected to the window, computed to fixed point.
Variables in $\mathcal{C}_\mathbf{w}\setminus\mathbf{w}$ are fixed
at iSAM2 MAP values, keeping $d=9$.

Likelihood evaluates nonlinear GTSAM factor error:
\begin{equation}
  \log\mathcal{L}(\Theta_\mathbf{w}^{(i)}) =
    -\tfrac{1}{2}\sum_{f\in\mathcal{F}_\mathbf{w}}
    f.\text{error}\!\left(\Theta_\mathbf{w}^{(i)}\cup
    \hat\Theta_{\mathcal{C}\setminus\mathbf{w}}\right)
  \label{eq:ll}
\end{equation}

Prior maps $[0,1]^d$ to $\mathcal{N}(\hat\Theta_\mathbf{w},
c\,\Sigma_\mathbf{w})$ via Cholesky and probit transform.

From weighted samples, posterior mean
$\tilde\Theta_\mathbf{w}=\sum_i w^{(i)}\Theta^{(i)}$ is computed
and applied if:
\begin{equation}
  \log p(\tilde\Theta_\mathbf{w}|\mathcal{Z}) \geq
  \log p(\hat\Theta_\mathbf{w}|\mathcal{Z}) - 10^{-3}
  \label{eq:gate}
\end{equation}
Passing windows are applied simultaneously.
Algorithm~\ref{alg:sngr} presents the full SNGR pipeline.

\begin{algorithm}[t]
\caption{Selective Non-Gaussian Refinement (SNGR)}
\label{alg:sngr}
\begin{algorithmic}[1]
\Require $\mathcal{G}$, $\mathcal{Z}$, $\tau$, $W$, $c$
\Ensure $\hat{\Theta}$
\For{each pose $t$}
  \State $\hat{\Theta}\gets\text{iSAM2.update}(\mathcal{G},\mathcal{Z}_t)$
\EndFor
\State $\text{pending}\gets\emptyset$
\For{each $\mathbf{w}=\{t,\ldots,t+W-1\}$, $t\geq W-1$}
  \State $s_\mathbf{w}\gets\log_{10}(\lambda_{\max}/\lambda_{\min})$
         \eqref{eq:score}
  \If{$s_\mathbf{w}>\tau$}
    \State compute $\mathcal{C}_\mathbf{w}$, $\mathcal{F}_\mathbf{w}$
    \State $\{(\Theta^{(i)},w^{(i)})\}\gets
           \text{NestedSampling}(\mathcal{F}_\mathbf{w},\pi,\ell)$
           \eqref{eq:ll}
    \State $\tilde{\Theta}_\mathbf{w}\gets\sum_i w^{(i)}\Theta^{(i)}$
    \If{gate \eqref{eq:gate} holds}
      \State $\text{pending}\gets\text{pending}\cup
             \{(\mathbf{w},\tilde{\Theta}_\mathbf{w})\}$
    \EndIf
  \EndIf
\EndFor
\State apply pending updates to $\hat{\Theta}$
\State \Return $\hat{\Theta}$
\end{algorithmic}
\end{algorithm}

\section{Implementation}

\subsection{Trigger-Only Baseline} 
To isolate trigger from refinement quality, we implement a baseline trigger-only pipeline
with a no-op rejection sampler: 500 samples from
$\mathcal{N}(\hat\Theta_\mathbf{w},0.5^2 I)$ weighted under the
iSAM2 marginal.
Since both are Gaussians at the same MAP, no refinement occurs.
Log importance weights average $-73$\,nats at $p=0.3$, confirming
the no-op character quantitatively.

\subsection{Nested Sampling SNGR Configuration}

The full pipeline uses dynesty~\cite{speagle2020dynesty} with
\texttt{rwalk} sampler, $n_\text{live}=100$,
$n_\text{walks}=\max(d,9)$, and $\Delta\log Z=0.5$ convergence.
\texttt{rwalk} handles non-ellipsoidal likelihoods without
enlargement warnings.

\subsection{Software and Hardware}

Experiments use GTSAM~4.2.0, dynesty~3.0.0, Python~3.11, running on
Apple M1 with 8\,GB unified memory.

\section{Benchmarking}
\subsection{Experimental Design}
 
SNGR is evaluated on a synthetic 2-D range-only SLAM scenario.
We choose a synthetic setting deliberately: it provides exact ground
truth for all poses and landmarks, allows precise control over the
type and degree of failure injected, and makes the evaluation
fully reproducible across seeds without dependence on a specific
real-world dataset or sensor platform.

 \begin{figure}[t]
  \centering
  \begin{subfigure}[b]{0.48\columnwidth}
    \includegraphics[width=\linewidth]{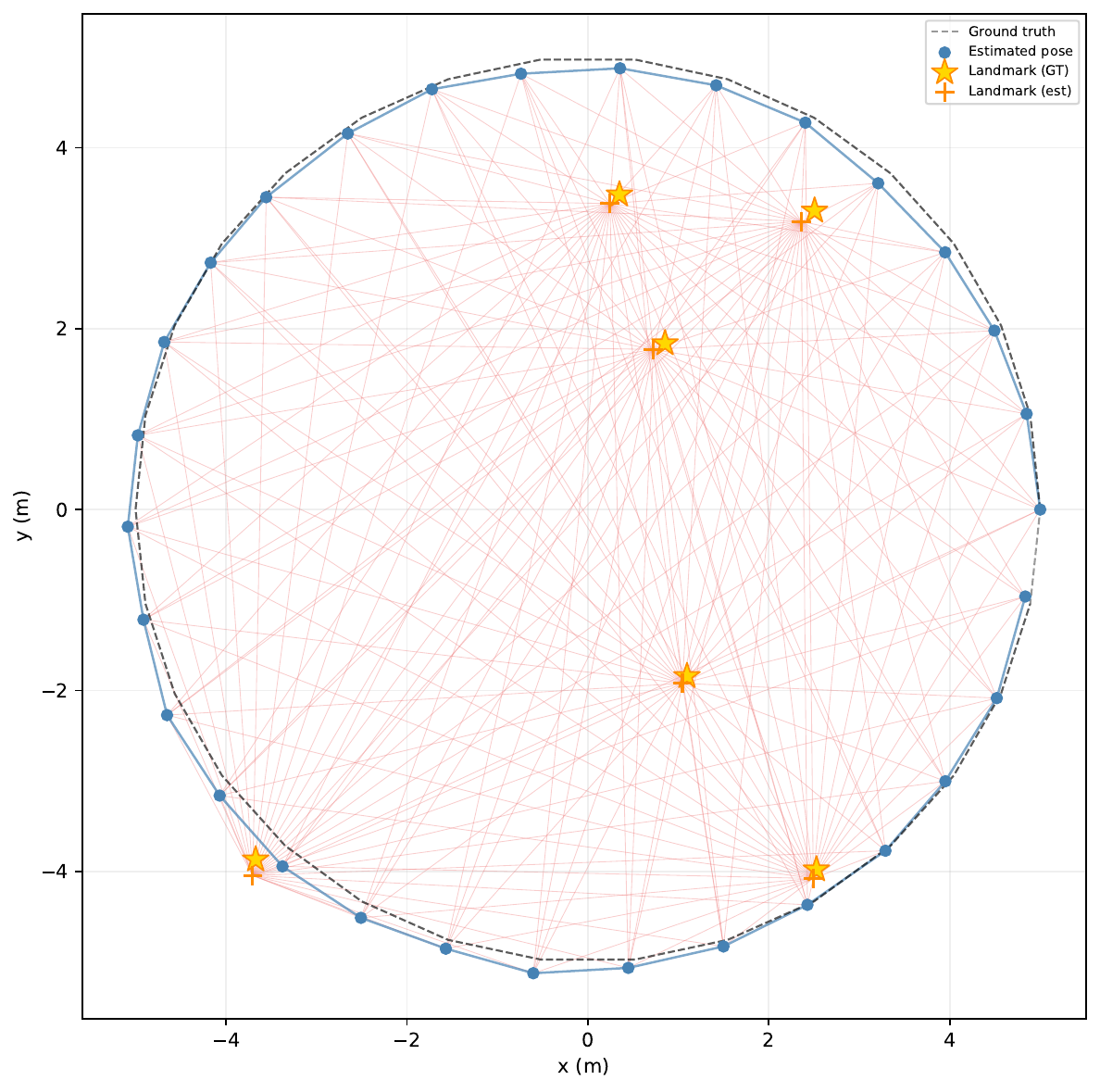}
    \caption{$p=0.0$, seed 0.
    The estimated trajectory closely tracks the ground truth.
    No windows trigger.}
  \end{subfigure}\hfill
  \begin{subfigure}[b]{0.48\columnwidth}
    \includegraphics[width=\linewidth]{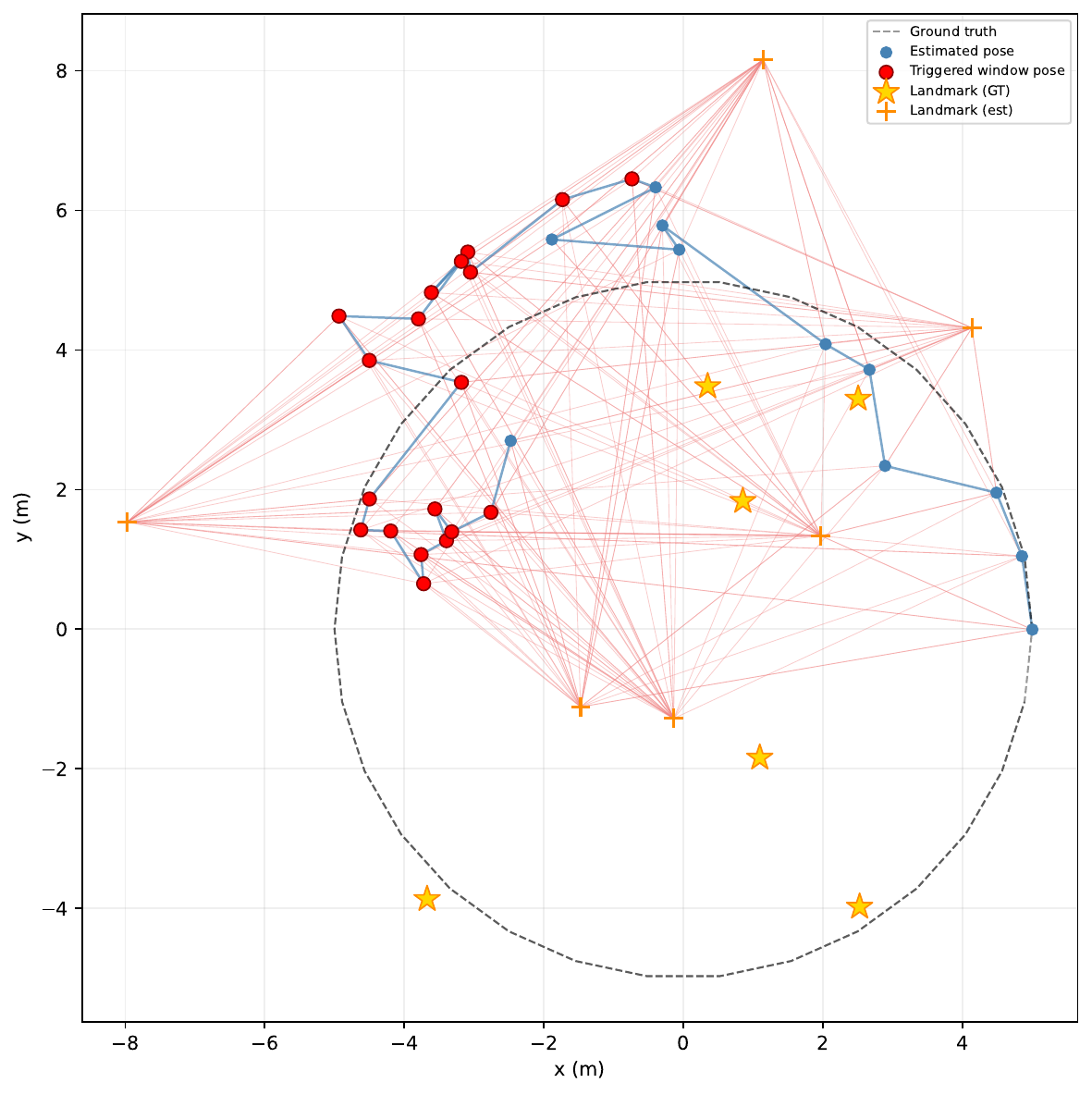}
    \caption{$p=0.3$, seed 0.
    Red poses are triggered windows.
    Long cross-graph factor edges reveal that poses are associated
    with geometrically distant wrong landmarks.}
  \end{subfigure}
  \caption{Factor graph visualization at two noise levels
  ($\tau=3.96$).
  Gold stars: ground-truth landmarks.
  Orange crosses: estimated landmarks.
  The visual difference in factor graph density between clean and
  noisy cases corresponds directly to the difference in condition
  number scores seen in Fig.~\ref{fig:scores}.}
  \label{fig:fg}
\end{figure}

The robot follows a circular trajectory of radius $5.0$\,m sampled
at $T=30$ uniformly spaced poses.
A circular path is chosen because it is a canonical stress test for
range-only SLAM: the robot observes each landmark from a continuously
changing direction, which is the geometry most likely to produce
well-conditioned range constraints under clean data, and therefore
provides a fair baseline against which failures introduced by
wrong association stand out clearly.
Six landmarks are drawn uniformly at random from $[-4,4]^2$\,m and
all six are observed from every pose, ensuring the factor graph is
fully connected and that any degradation in estimate quality is
attributable to the injected noise rather than sparse coverage.
 
Odometry is corrupted by additive Gaussian noise with translational
standard deviation $\sigma_t=0.05$\,m and rotational standard
deviation $\sigma_r=0.01$\,rad per step, representing a moderately
accurate odometer where drift accumulates slowly enough that
range corrections are meaningful.
Range measurements are corrupted by zero-mean Gaussian noise with
$\sigma_{\text{range}}=0.1$\,m, consistent with a short-range
ultrasonic or UWB ranging sensor.
 
Wrong-association noise is the primary experimental variable.
It is applied independently to each measurement with probability
$p_\text{noise}$: a corrupted measurement replaces the true landmark
index $k$ with an index $j \neq k$ drawn uniformly at random from
the remaining $K-1$ landmarks, producing a factor that is
geometrically irreconcilable with the true trajectory.
This models the failure mode where a data association front-end
produces hard assignments without uncertainty, a common assumption
in practical SLAM pipelines.
Five noise levels $p_\text{noise}\in\{0.0,0.1,0.2,0.3,0.4\}$ are
evaluated, spanning the range from a clean baseline through moderate
contamination to a heavily corrupted regime where nearly half of all
measurements are wrong.
Each level is evaluated across five random seeds to account for
variability in landmark geometry, which we show has a material effect
on trigger behaviour.
RMSE is reported as mean~$\pm$~std across seeds; NEES, trigger
counts, $\Delta\!\log p$, and timing are reported from seed~0 for
conciseness.
 
\subsection{Evaluation Strategy}
 
The benchmarking is structured in three stages, each isolating a
different component of SNGR.
 
The first stage is a \textit{trigger evaluation} using the
trigger-only baseline, which runs the full condition-number detection
pipeline but replaces the nested sampler with a no-op importance
sampler.
This isolates trigger precision and recall from any effect of the
refinement quality, allowing us to characterise the detector
independently.
 
The second stage is the \textit{full SNGR pipeline evaluation},
which adds dynesty nested sampling to every triggered window and
applies the gating rule.
Comparing RMSE and NEES between this and iSAM2 alone measures
whether local refinement translates into global trajectory
improvement, and $\Delta\!\log p$ quantifies the quality of the
local posterior correction regardless of global effect.
 
The third stage is the \textit{bimodal proof-of-concept}, a
deliberately minimal two-anchor scenario that tests whether nested
sampling can recover a bimodal posterior that iSAM2 is structurally
incapable of representing.
This is evaluated analytically rather than statistically since the
ground truth is known exactly.
 
\subsection{Metrics}
 
Four complementary metrics are used, each targeting a different
aspect of the system.
 
\textbf{RMSE}~\cite{sturm2012,zhang2018traj} measures global
trajectory accuracy:
$\sqrt{\tfrac{1}{T}\sum_t\left\|\hat{x}_t^{[x,y]}-x_t^\text{gt}
\right\|_2^2}$.
It is the primary measure of whether SNGR improves the estimate
that a downstream planner or mapper would consume.
 
\textbf{NEES}~\cite{barshalom2001} measures estimator consistency,
i.e.\ whether the reported covariance matches the actual error:
$(x_t-\hat{x}_t)^\top\mathbf{P}_t^{-1}(x_t-\hat{x}_t)$,
where $\mathbf{P}_t$ is the $2\times2$ marginal position covariance.
For a consistent estimator this quantity follows $\chi^2_2$ with
mean~2; values far above 2 indicate overconfidence.
NEES is critical here because wrong-association failures often
produce small covariances alongside large errors, which
RMSE alone cannot characterize.
 
\textbf{Trigger precision and recall} against ground-truth failure
windows $\Phi$~\eqref{eq:fail} measure the detection quality of the
condition-number trigger independently of the sampler.
Precision quantifies false-positive rate (wasted sampling cost);
recall quantifies coverage of true failures.
 
\textbf{$\Delta\!\log p$} measures the improvement in local
posterior log-likelihood of the nested sampling mean over the iSAM2
MAP on the triggered window's factor subgraph.
A positive value confirms the sampler found a genuinely better
local solution; a value near zero confirms the gate is correctly
suppressing no-op updates on false-positive triggers.
 
\textbf{Wall-clock time} is reported separately for iSAM2 processing
and for nested sampling refinement, and is compared against the
hypothetical cost of running nested sampling on all 28 windows
exhaustively.

\section{Results}

\subsection{Bimodal Experiment}
\label{sec:bimodal}

\begin{figure}[!t]
  \centering
  \begin{subfigure}[b]{0.48\columnwidth}
    \includegraphics[width=\linewidth]{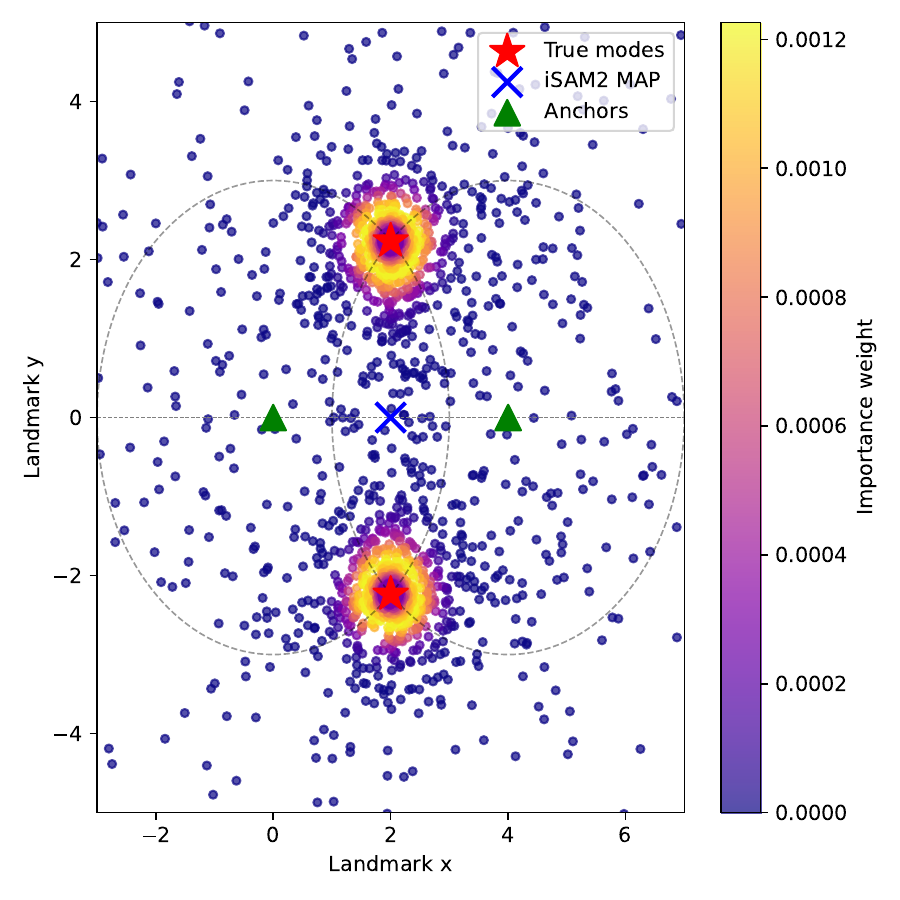}
    \caption{Weighted samples over $L(0)$.
    iSAM2 MAP (blue $\times$) at saddle $(2,0)$; true modes
    (red $\bigstar$) receive equal mass.}
  \end{subfigure}\hfill
  \begin{subfigure}[b]{0.48\columnwidth}
    \includegraphics[width=\linewidth]{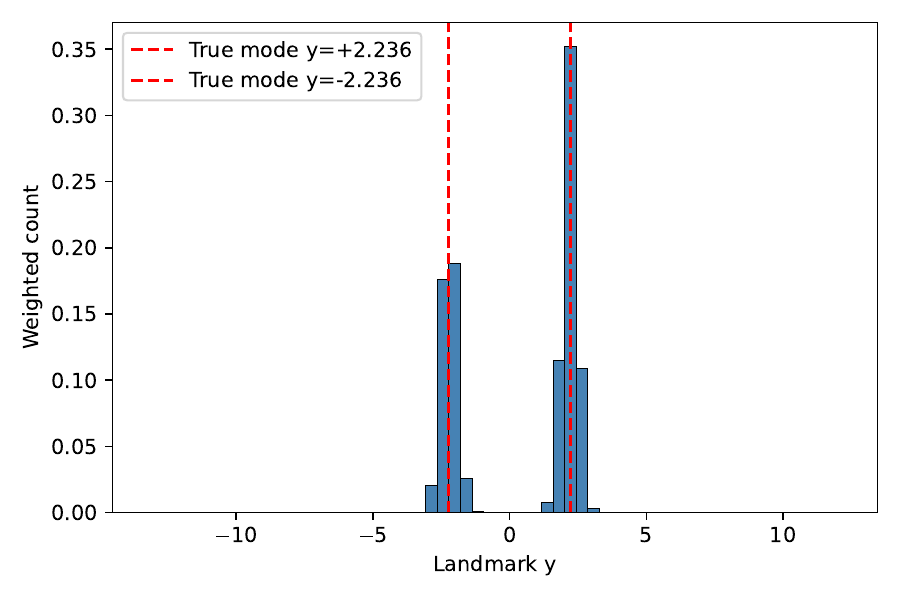}
    \caption{Marginal histogram of $L(0)_y$.
    Two peaks at $y\approx\pm2.236$; MAP at $y=0$ receives
    negligible weight.}
  \end{subfigure}
  \caption{Bimodal proof-of-concept. Anchors at $A=(0,0)$, $B=(4,0)$
  observe $L(0)$ at $r=3.0$\,m. Constraint manifold bimodal at
  $(2,\pm\sqrt{5})$. iSAM2 converges to saddle; nested sampling
  recovers both modes.}
  \label{fig:bimodal}
\end{figure}

\begin{table}[!t]
  \caption{Bimodal Experiment: iSAM2 vs Nested Sampling}
  \label{tab:bimodal}
  \centering
  \begin{tabular}{lcc}
    \toprule
    & iSAM2 & Nested Sampling \\
    \midrule
    $L(0)$ estimate (m)      & $(2.000,0.000)$ & $(1.984,-2.232)$ \\
    Log-likelihood (nats)    & $-12.50$        & $-0.01$ \\
    Error from mode (m)      & 2.236           & 0.004 \\
    $\Delta\!\log p$         & ---             & $+12.49$ \\
    Weighted $\sigma_y$ (m)  & ---             & $2.21$ \\
    Bimodality coefficient   & ---             & $0.62$ \\
    \bottomrule
  \end{tabular}
\end{table}

Two anchors at $A=(0,0)$, $B=(4,0)$ each observe $L(0)$ at $r=3.0$\,m.
The constraint manifold is
$\{L:\left\|L-A\right\|_2=3\}\cap\{L:\left\|L-B\right\|_2=3\}
=\{(2,\pm\sqrt{5})\}$.

iSAM2 initializes at $(2,0)$ and converges there by gradient
cancellation: range factor gradients from $A$ and $B$ point in
opposite directions.
This saddle is 12.49\,nats below either mode.

Nested sampling ($c=50$) discovers both modes.
Best sample at $(1.984,-2.232)$, within 0.004\,m of true mode, with
$\Delta\!\log p=+12.49$\,nats.
Weighted $\sigma_y=2.21$\,m confirms equal mass on both modes (true:
$\sqrt{5}=2.24$\,m).
Bimodality coefficient 0.62 exceeds the 0.555
threshold~\cite{freeman2013}.

\subsection{Trigger Analysis}
\label{sec:trigger}

\begin{figure}[!t]
  \centering
  \begin{subfigure}[b]{0.48\columnwidth}
    \includegraphics[width=\linewidth]{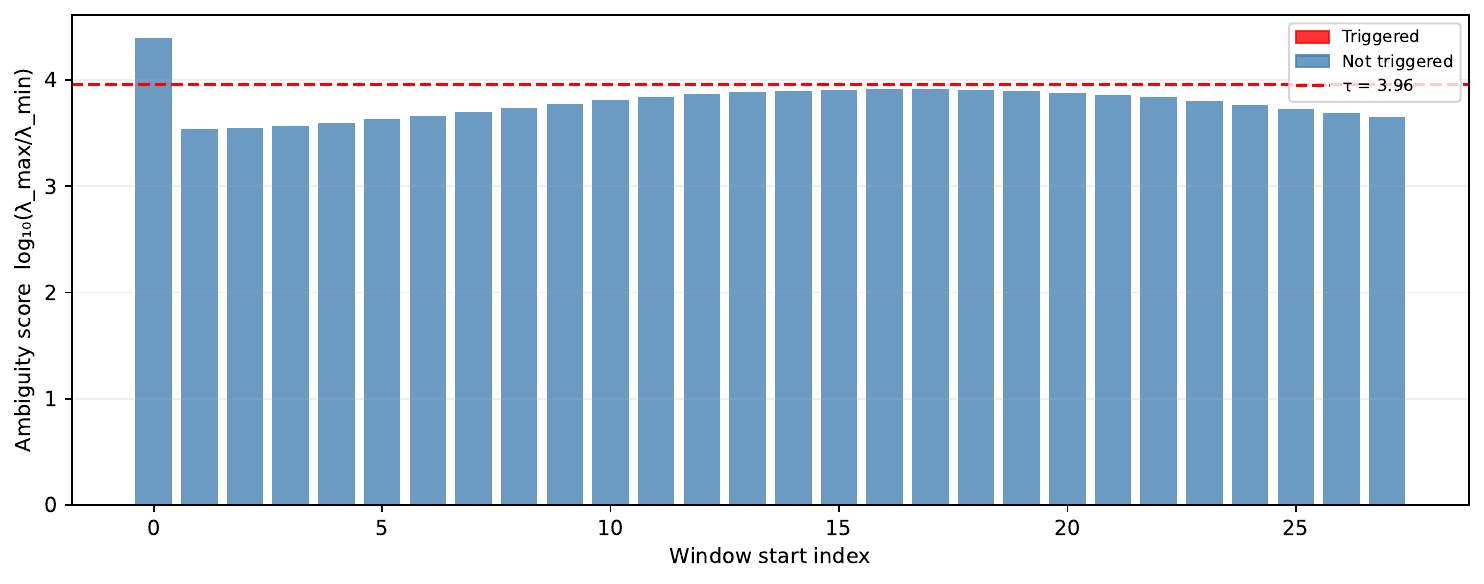}
    \caption{$p=0.0$: all windows score below $\tau=3.96$.}
  \end{subfigure}\hfill
  \begin{subfigure}[b]{0.48\columnwidth}
    \includegraphics[width=\linewidth]{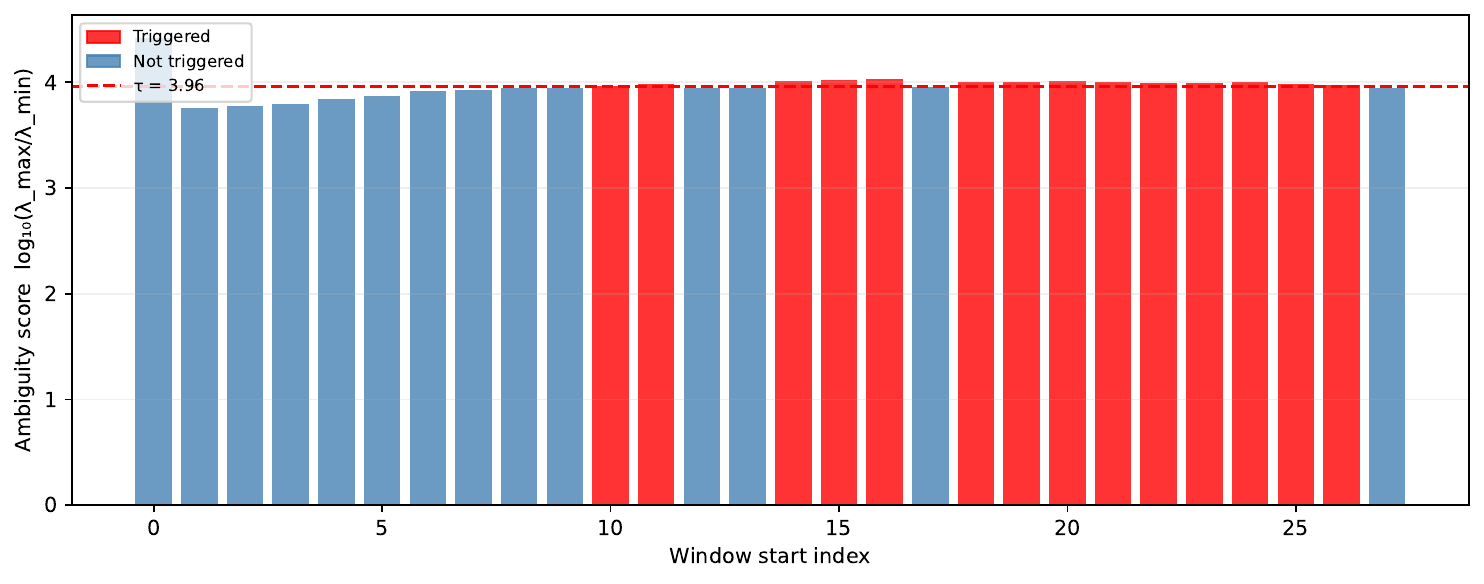}
    \caption{$p=0.3$: most windows exceed threshold.}
  \end{subfigure}
  \caption{Per-window scores $s_\mathbf{w}$ (seed~0, $\tau=3.96$).
  Distributions at $p=0.0$ and $p=0.1$ are identical, confirming
  the blind spot is structural.}
  \label{fig:scores}
\end{figure}

\begin{figure}[!t]
  \centering
  \begin{subfigure}[b]{0.48\columnwidth}
    \includegraphics[width=\linewidth]{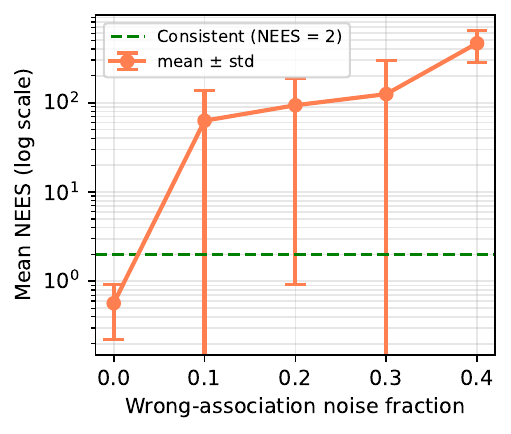}
    \caption{NEES (log scale) vs noise. At $p=0.1$, NEES$=148$
    with zero triggers.}
  \end{subfigure}\hfill
  \begin{subfigure}[b]{0.48\columnwidth}
    \includegraphics[width=\linewidth]{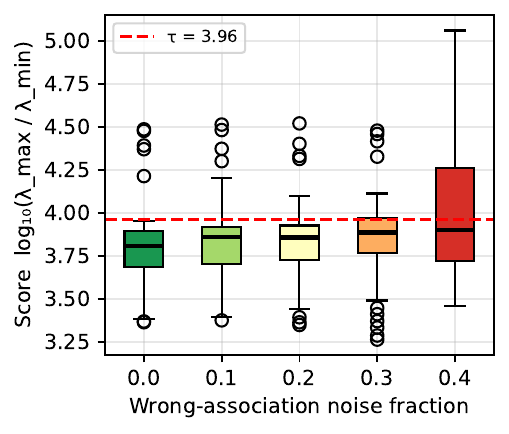}
    \caption{Score distributions per noise level. $p=0.0$ and
    $p=0.1$ boxes overlap.}
  \end{subfigure}
  \caption{Trigger-only baseline diagnostics ($\tau=3.96$, 5 seeds).
  Overconfidence grows with noise; trigger remains blind at low
  contamination.}
  \label{fig:baseline_fig}
\end{figure}

\begin{table}[!t]
  \caption{Threshold Sensitivity (Seed~0)}
  \label{tab:tau}
  \centering
  \begin{tabular}{ccc}
    \toprule
    $\tau$ & Clean & Noisy ($p=0.3$) \\
    \midrule
    3.92 &  0/28 & 21/28 \\
    3.96 &  0/28 & 14/28 \\
    4.0  &  0/28 &  8/28 \\
    \bottomrule
  \end{tabular}
\end{table}

\begin{table}[!t]
  \caption{Baseline Results (5-Seed RMSE; Seed~0 NEES, $\tau=3.96$)}
  \label{tab:baseline}
  \centering
  \setlength{\tabcolsep}{3.5pt}
  \begin{tabular}{cccccc}
    \toprule
    $p$ & RMSE (m) & NEES & Prec. & Recall & Trig./Fail \\
    \midrule
    0.0 & $0.15\pm0.06$ &   0.36 & --- & ---
        & 0/0 \\
    0.1 & $1.27\pm1.20$ & 148.52 & --- & 0.00
        & 0/25 \\
    0.2 & $1.64\pm0.90$ & 210.15 & 1.00 & 0.15
        & 4/26 \\
    0.3 & $1.82\pm1.21$ & 458.26 & 1.00 & 0.54
        & 14/26 \\
    0.4 & $4.14\pm1.24$ & 745.23 & 0.96 & 0.89
        & 24/26 \\
    \bottomrule
  \end{tabular}
\end{table}

\begin{table}[!t]
  \caption{Per-Seed Breakdown at $p=0.3$, $\tau=3.96$}
  \label{tab:perseed}
  \centering
  \begin{tabular}{ccccccc}
    \toprule
    Seed & Trig. & TP & FP & FN & Prec. & Recall \\
    \midrule
    0 & 14/28 & 14 & 0 & 12 & 1.00 & 0.54 \\
    1 &  8/28 &  8 & 0 & 16 & 1.00 & 0.33 \\
    2 &  0/28 &  0 & 0 & 17 & ---  & 0.00 \\
    3 & 10/28 & 10 & 2 & 13 & 0.83 & 0.44 \\
    4 &  0/28 &  0 & 0 & 20 & ---  & 0.00 \\
    \bottomrule
  \end{tabular}
\end{table}

Tables~\ref{tab:tau}--\ref{tab:perseed} and
Figs.~\ref{fig:scores}--\ref{fig:baseline_fig} summarize baseline
results.

At $p=0.1$, NEES $=148$ with zero triggers: covariance shape and
accuracy are independent properties.
At 10\% contamination, corrupted factors spread uniformly and their
MAP displacements partially cancel, leaving isotropic covariance
despite trajectory bias.

At $p\geq0.2$, enough corruption accumulates to produce elongation
and the trigger fires with precision~$=1.0$.
Table~\ref{tab:perseed} shows seeds 2 and 4 produce zero triggers
despite 17--20 failures; landmark geometry keeps displacements
isotropic in these instances.

\subsection{Full SNGR Pipeline}
\label{sec:full}

\begin{table}[!t]
  \caption{SNGR Results (Seed~0, $\tau=3.9$)}
  \label{tab:sngr}
  \centering
  \setlength{\tabcolsep}{4pt}
  \begin{tabular}{ccccccc}
    \toprule
    $p$ & Pose & LM & NEES & Trig. & Prec. & Recall \\
    \midrule
    0.0 & 0.13 & 0.10 &   0.36 &  4/28 & 0.00 & --- \\
    0.1 & 2.81 & 3.31 & 148.52 &  0/28 & ---  & 0.00 \\
    0.2 & 2.65 & 2.30 & 209.88 & 14/28 & 1.00 & 0.54 \\
    0.3 & 3.81 & 2.91 & 454.14 & 22/28 & 1.00 & 0.85 \\
    \bottomrule
  \end{tabular}
\end{table}

\begin{table}[!t]
  \caption{Cost: Selective vs Exhaustive (Seed~0)}
  \label{tab:cost}
  \centering
  \begin{tabular}{cccccc}
    \toprule
    $p$ & Trig. & iSAM2 & Refine & Exhaust. & Saving \\
    \midrule
    0.0 &  4/28 & 0.02\,s &  44\,s & 336\,s & $7.6\times$ \\
    0.1 &  0/28 & 0.02\,s &   0\,s & 336\,s & --- \\
    0.2 & 14/28 & 0.02\,s & 175\,s & 336\,s & $1.9\times$ \\
    0.3 & 22/28 & 0.02\,s & 245\,s & 336\,s & $1.4\times$ \\
    \bottomrule
  \end{tabular}
\end{table}

\begin{figure}[!t]
  \centering
  \begin{subfigure}[b]{0.48\columnwidth}
    \includegraphics[width=\linewidth]{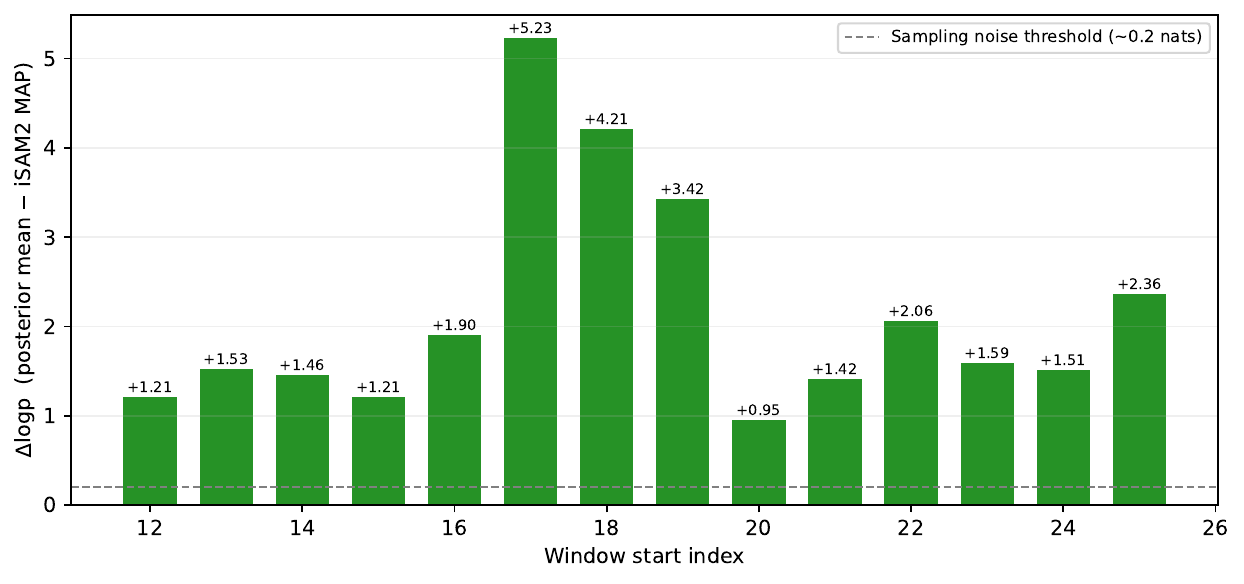}
    \caption{$\Delta\!\log p$ per window, $p=0.2$, seed~0.
    All values exceed 0.2\,nat noise floor.
    Peak $+5.04$\,nats at window~17.}
  \end{subfigure}\hfill
  \begin{subfigure}[b]{0.48\columnwidth}
    \includegraphics[width=\linewidth]{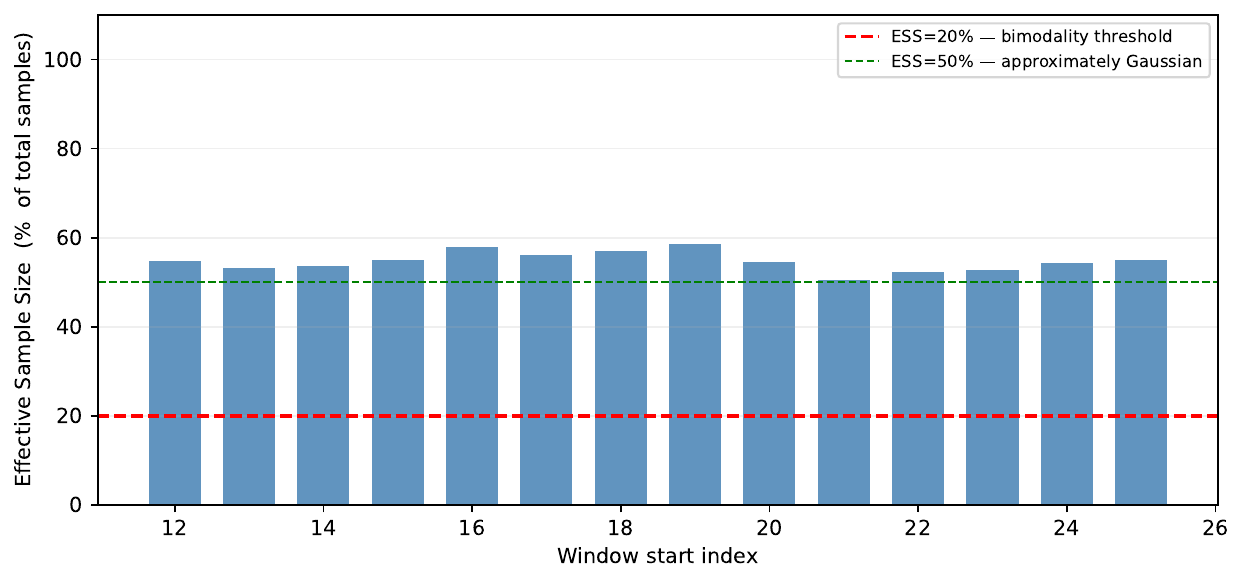}
    \caption{ESS per window, $p=0.2$, seed~0.
    Values near 50\% confirm Gaussian local posteriors.}
  \end{subfigure}
  \caption{SNGR diagnostics at $p=0.2$. All improvements genuine;
  ESS indicates within-mode corrections.}
  \label{fig:diagnostics}
\end{figure}

Table~\ref{tab:sngr} and Fig.~\ref{fig:diagnostics} show full
pipeline results.

At $p=0.0$, four windows trigger (false positives at $\tau=3.9$).
ESS~$\approx50\%$ and $\Delta\!\log p\approx0$; the gate retains
MAP in all cases.

At $p=0.1$, the blind spot persists: zero triggers despite
NEES~$=148$.

At $p=0.2$, all 14 windows show $\Delta\!\log p>0$ ($+1.0$ to
$+5.04$\,nats), exceeding the noise floor.
ESS near 50\% indicates Gaussian local posteriors; the sampler finds
within-mode corrections.

Despite $\Delta\!\log p>0$ on every window, global RMSE does not
improve.
Wrong-association errors distribute throughout the odometry chain;
correcting a window leaves surrounding odometry factors unchanged.
This identifies SNGR's scope: effective when failures are localized,
not when contamination is globally distributed.

Per-window cost is constant at 11--12\,s (Table~\ref{tab:cost}).
Selective triggering provides $7.6\times$ savings at $p=0.0$,
$1.9\times$ at $p=0.2$.
At $p=0.3$, 22/28 windows trigger and selectivity is lost.

\FloatBarrier
\section{Conclusion and Future Work}

We presented Selective Non-Gaussian Refinement, a method that wraps
iSAM2 with a triggered nested sampling layer to address the specific
failure modes where Gaussian posterior approximations break down by
construction.
By exploiting the marginal covariance structure available from the
iSAM2 Bayes tree, SNGR applies non-Gaussian inference selectively
on the windows where it is most likely to help rather than uniformly
across the entire trajectory.
The condition-number trigger achieves precision $=1.0$ above 20\%
wrong-association contamination with no false positives on clean data.
The nested sampler finds genuine local improvements of $+1.0$ to
$+5.842$\,nats per triggered window at $p=0.2$, and the bimodal
experiment yields a $+12.49$\,nat improvement in a setting where
iSAM2 is structurally incapable of finding the correct answer.
Selective triggering provides $7.6\times$ cost savings at moderate
noise; per-window cost is constant at 11 to 12\,s regardless of
noise level.

The $p=0.1$ results are as informative as the positive ones.
NEES $=148$ with zero triggers demonstrates concretely that covariance
shape and covariance accuracy are independent properties, and that a
shape-based detector cannot see all the failures that matter.
The global RMSE finding shows that windowed local refinement is
orthogonal to trajectory drift transmitted through the odometry
chain, and is therefore appropriate only when failures are spatially
isolated.
SNGR is effective when failures are geometrically localised and
produce detectable covariance elongation; it does not detect
isotropically distributed wrong-association noise and cannot correct
globally distributed trajectory drift.

Three specific extensions address the current limitations of SNGR.

\textbf{Residual-based complementary trigger.}
Since the condition-number trigger cannot detect failures that do not
produce covariance elongation, the natural complement is a per-window
residual test.
The mean squared normalized residual
$\bar{r}^2_\mathbf{w} = |\mathcal{F}_\mathbf{w}|^{-1}
\sum_f (z_f - h_f(\hat\Theta))^2 / \sigma_f^2$
follows a $\chi^2$ distribution under the correct Gaussian model, and
a value significantly exceeding its expected value identifies windows
where the Gaussian model is a poor fit, regardless of whether the
covariance is elongated or isotropic.
Combining both criteria with a logical OR would close the blind spot
at low contamination rates while preserving precision at higher noise.

\textbf{Principled threshold calibration.}
The current threshold $\tau$ is tuned on the same scenario used for
evaluation, which is a limitation.
The 0.08-unit operating range makes it fragile to changes in
trajectory geometry, landmark density, or sensor parameters.
A more robust approach would derive $\tau$ analytically from the null
distribution of $s_\mathbf{w}$ under correct Gaussian assumptions,
or estimate this distribution from a short clean-data calibration
run, yielding a threshold with a controlled false-positive rate that
transfers across scenarios.

\textbf{Ancestral prior sampling for distant modes.}
With $c = 1.0$, the Gaussian prior restricts the sampler to the MAP
neighborhood.
The ESS near 50\% observed across all triggered windows at $p=0.2$
confirms that the sampler is finding within-mode corrections, not
discovering topologically distant alternative hypotheses.
For loop-closure scenarios where alternative trajectory hypotheses
are far from the current MAP, the ancestral prior sampling approach
of NSFG~\cite{huang2022nested} would provide coverage of the full
prior volume without requiring manual tuning of the inflation
parameter $c$.
Integrating this into the SNGR framework and evaluating on real-world
datasets with genuine loop-closure ambiguity is the most direct path
toward handling multimodal posteriors in an online selective setting.

SNGR in its current form is most useful as an offline accuracy
evaluation tool for approximate SLAM posteriors and as a building
block for more complete selective non-Gaussian inference pipelines.

\section*{Acknowledgment}
The authors thank Prof. S. Hutchinson for guidance on factor graph
SLAM and M. Potter for support.

\bibliographystyle{IEEEtran}
\bibliography{references}

\end{document}